\pdfoutput=1

\documentclass[11pt]{article}

\usepackage[]{naacl2021}

\usepackage{times}
\usepackage{latexsym}
\usepackage{booktabs}
\usepackage{arydshln}
\usepackage{graphicx}
\usepackage{amsmath}

\usepackage[T1]{fontenc}

\usepackage[utf8]{inputenc}

\usepackage{microtype}
\usepackage{graphics}
\usepackage{tablefootnote}

\title{Revisiting Simple Neural Probabilistic Language Models}

\author{Simeng Sun \and Mohit Iyyer \\
        College of Information and Computer Sciences\\ University of Massachusetts Amherst \\
        \texttt{\{simengsun, miyyer\}@cs.umass.edu}}

\begin{document}
\maketitle

\newcommand\wtthree{\textsc{Wikitext-103}}
\newcommand\wttwo{\textsc{Wikitext-2}}
\newcommand\lambada{\textsc{lambada}}
\newcommand\enwik{\textsc{enwik8}}
\newcommand{\bmat}[1]{\text{\textbf{#1}}}
\newcommand{\bvec}[1]{\boldsymbol{#1}}

\begin{abstract}
Recent progress in language modeling has been driven not only by advances in neural architectures, but also through hardware and optimization improvements. In this paper, we revisit the neural probabilistic language model (NPLM) of~\citet{Bengio2003ANP}, which simply concatenates word embeddings within a fixed window and passes the result through a feed-forward network to predict the next word. When scaled up to modern hardware, this model (despite its many limitations) performs much better than expected on word-level language model benchmarks. Our analysis reveals that the NPLM achieves lower perplexity than a baseline Transformer with short input contexts but struggles to handle long-term dependencies. Inspired by this result, we modify the Transformer by replacing its first self-attention layer with the NPLM's local concatenation layer, which results in small but consistent perplexity decreases across three word-level language modeling datasets.  
\end{abstract}


\section{Introduction}

Over the past decade, state-of-the-art neural architectures for language modeling (LM) have transitioned from simple recurrent neural networks~\citep{mikolov2011extensions} to LSTMs~\citep{zaremba2014recurrent} and finally to Transformers~\citep{NIPS2017_3f5ee243}. This progress is not due solely to LM-specific advances, however, as general-purpose upgrades such as residual connections~\citep{he2016deep} and layer normalization~\citep{ba2016layer} have enabled scaling to huge datasets and model sizes~\citep{kaplan2020scaling} on powerful GPUs.

In this paper, we revisit the neural probabilistic language model (NPLM) of~\citet{Bengio2003ANP}, the first (and simplest) neural architecture proposed for language modeling, through the lens of modern architecture design, hardware, and optimization. Given an input sequence of tokens, the NPLM first concatenates the previous $n$ token embeddings and then passes the result through a feed-forward network to predict the next token. Due to its small context window and lack of parameter sharing, the NPLM has been rendered obsolete, discarded in favor of LSTMs and Transformers.

To what extent are its limitations mitigated by modern design and optimization choices?
To answer this question,  we design an upgraded NPLM featuring increased depth and window size $n$ that incorporates residual connections, layer normalization, and dropout. We also include global context representations  to the concatenation layer by applying simple aggregation functions to embeddings outside of the local context window. These modifications substantially improve the NPLM: on the \wtthree\ benchmark dataset, the original NPLM of~\citet{Bengio2003ANP} reaches a validation perplexity of \textbf{216}, compared to \textbf{31.7} for our implementation, and \textbf{25.0} for a Transformer baseline.  

\begin{figure}[t]
    \includegraphics[width=0.45\textwidth]{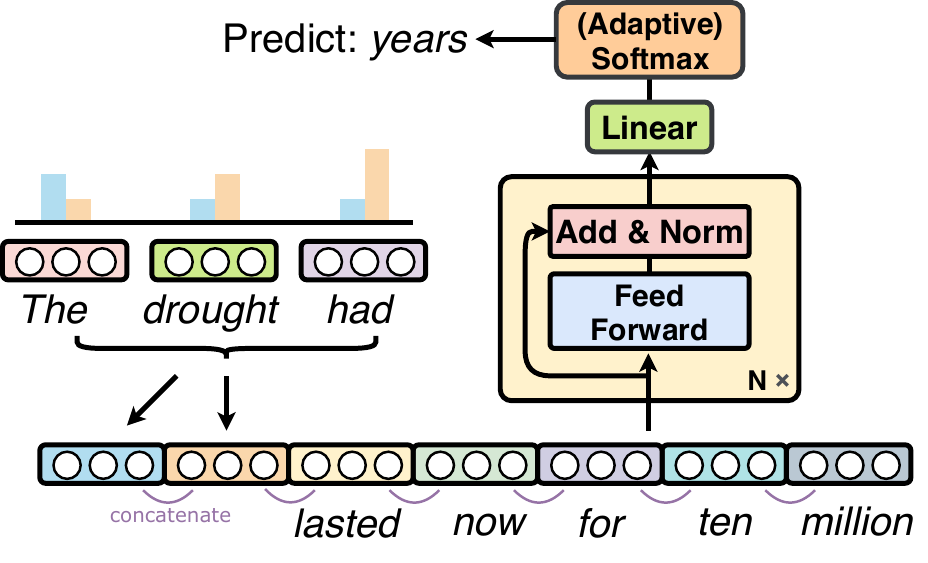}
    \caption{A modernized version of the neural probabilistic language model of~\citet{Bengio2003ANP}, which concatenates token embeddings within a fixed local window and feeds them to a stack of feed-forward layers to predict the next token. Our modified version additionally concatenates representations of the  distant context, which are computed by applying a weighted average to token representations outside the local window.}
    \label{fig:nplm}
\end{figure}

Can we improve Transformer language models by hybridizing them with NPLMs? Interestingly, we discover that our NPLM actually \emph{outperforms} the Transformer when given shorter input contexts (Figure~\ref{fig:short-context-l0-constrain}), although it is unable to take full advantage of longer contexts. Inspired by this result, we create two simple variants of the Transformer, one in which the first self-attention layer is replaced with the NPLM's concatenation layer, and the other in which self-attention in the first layer is constrained to a small local window.\footnote{Code available at \url{https://github.com/SimengSun/revisit-nplm}} These adjustments result in small but consistent perplexity decreases compared to a baseline Transformer across three word-level language modeling datasets (the first variant obtains \textbf{24.1} validation perplexity on \wtthree). Our qualitative analysis shows that the modified Transformers are better at predicting rare tokens and named entities, especially those that have already appeared in the context.

\section{Neural probabilistic language models}

Modern neural language models (NLMs) compute the conditional probability of a token $w_t$ given preceding (or \emph{prefix}) tokens $w_{<t}$ by first computing a dense vector representation of the prefix and then feeding it into a classifier to predict the next word. More concretely, a composition function $g$ is applied to the sequence of token embeddings $\bvec{x}_{<t}$ associated with the prefix, which results in a dense vector $\bvec{z} = g(\bvec{x}_{<t})$. A softmax classifier then takes $\bvec{z}$ as input and produces a distribution $P(w_t \mid w_{<t})$ over the vocabulary. Transformers~\citep{NIPS2017_3f5ee243} are currently the most popular choice for the composition function $g$.


\paragraph{NPLM definition:} First introduced by~\citet{Bengio2003ANP}, the NPLM uses a simple composition function reminiscent of $n$-gram language modeling. It concatenates the last $k$ prefix embeddings and passes the result through a feed-forward layer:

\begin{equation}
    \bvec{z} = \tanh(\bmat{W}[\bvec{x}_{t-k-1}; \bvec{x}_{t-k} \dots ; \bvec{x}_{t-1}])
    \label{eq:concat}
\end{equation}

The NPLM has many intuitive limitations: (1) it ignores the global context provided by prefix tokens further than $k$ tokens away; (2) it uses a different set of parameters for each position in the prefix window; and (3) it has a relatively small number of parameters, which limits its expressivity.

\subsection{A modern update to the NPLM}
To what extent are these limitations mitigated after scaling up the NPLM using modern advances in neural network training? Here, we investigate the impact of a number of modifications to the NPLM on 
\wtthree\ validation perplexity (all results in Table~\ref{tab:ablation}).

\begin{table}[t]
    \centering
    \scalebox{0.79}{
    \begin{tabular}{lrr}
        \toprule
        \textbf{Model} & \textbf{\# Params} &\textbf{Val. perplexity} \\\hline
        Transformer & 148M & 25.0\\\hline
        NPLM-old & 32M\tablefootnote{Similar to~\cite{Bengio2003ANP} we set embedding dimension to 60 and hidden dimension to 100.} &  216.0 	  \\
        NPLM-old (large) & 221M\tablefootnote{We use the same embedding dimension and hidden dimension of our modern NPLM model. Weights are not tied.} &  128.2  \\\midrule
        NPLM 1L & 123M & 52.8  \\
        NPLM 4L & 128M & 38.3   \\
        NPLM 16L  & 148M & \textbf{31.7} \\ 
       \hspace{0.3cm} - Residual connections & 148M & 660.0 \\
       \hspace{0.3cm} - Adam, + SGD & 148M & 418.5 \\
       \hspace{0.3cm} - Global embedding & 146M & 41.9  \\
       \hspace{0.3cm} - Global kernel, + average & 148M & 37.7 \\
       \hspace{0.3cm} - Layer normalization & 148M & 33.0 \\
       \bottomrule
    \end{tabular}
    }
    \caption{NPLM model ablation on \wtthree.
    }
    \label{tab:ablation}
\end{table}

\paragraph{Increased depth and dimensionality:} We pass the concatenated representation into a multi-layer network instead of a single layer, and we also substantially increase the embedding and hidden layer dimensionality to 410 and 2100 respectively. \wtthree\ validation perplexity drops from \textbf{216} for the original one-layer NPLM (32M parameters) to \textbf{41.9} for a 16-layer NPLM with 148M parameters (no global prefix embeddings).

\paragraph{Better optimization for deep networks:} To improve gradient flow across the multi-layer network, we apply residual connections~\citep{he2016deep} and layer normalization~\citep{ba2016layer} at each layer. We additionally apply dropout~\citep{JMLR:v15:srivastava14a}, use rectified linear units (ReLU) instead of the $\tanh$ non-linearity, and train our NPLM with the Adam optimizer~\citep{kingma2015adam}.\footnote{Similar to ~\citet{baevski2018adaptive}, we first linearly warm up learning rate for 4K steps and then anneal with one cycle cosine learning rate scheduler. We did not observe improvements annealing with cyclical scheduler.} These modifications are \emph{crucial} for training our 16-layer NPLM: without residual connections, we reach a perplexity of \textbf{660}, while using standard SGD instead of Adam yields a perplexity of \textbf{418.5}.

\paragraph{Increased window size:} While hardware considerations limited the window size $k$ of the original NPLM to just five tokens, modern GPUs allow us to quickly train models with much larger memory footprints. We train models up to $k=50$ (Figure~\ref{fig:short-context-l0-constrain}) and observe perplexity drop from \textbf{87} with $k=3$ to eventually plateau around \textbf{40} with $k=50$. The plot also shows that Transformers take far better advantage of longer inputs.

\begin{table*}[t]
\centering
\scalebox{0.9}{
\begin{tabular}{@{}lccccllll@{}}
\toprule
& \multicolumn{2}{c}{\wttwo\ (13M)}       & \multicolumn{2}{c}{\wtthree\ (148M)}            & \multicolumn{2}{c}{\lambada\ (115M)}     & \multicolumn{2}{c}{\enwik\ (38M)}    \\ \cmidrule{2-3} \cmidrule{4-5} \cmidrule{6-7} \cmidrule{8-9}
& \multicolumn{1}{l}{Valid ppl.} & \multicolumn{1}{l}{Test ppl.} & \multicolumn{1}{l}{Valid ppl.} & \multicolumn{1}{l}{Test ppl.} & Valid ppl.                & Test ppl.                 & Valid bpc.                & Test bpc.                 \\\midrule
NPLM & 120.5 & 114.3 & 31.7 & 32.9 & 44.8 & 44.5 &  1.63 & 1.63\\
Transformer & 117.6 & 111.1 & 25.0 & 26.1 & 42.1 & 41.8 & \textbf{1.14} & \textbf{1.12}\\
Transformer-C & 113.1 & 107.5 & 24.1 & \textbf{25.1} & 42.0 & 41.7 & 1.14 & 1.12\\
Transformer-N & \textbf{110.8} & \textbf{105.6} & \textbf{24.1} & 25.2 & \textbf{41.8} & \textbf{41.5} & 1.14 & 1.12\\
\bottomrule
\end{tabular}
}
\caption{Our Transformer variants improve on the baseline Transformer across three word-level LM datasets. The \# of model parameters is shown in brackets (same for all models). For model details, see Appendix \ref{appendix:config}. }
    \label{tab:main-res}
\end{table*}

\paragraph{Tied weights and adaptive softmax:} The original NPLM computes probabilities of \emph{all} words in the vocabulary. For datasets with a large vocabulary, we use adaptive softmax~\cite{pmlr-v70-grave17a} to speed up training and decrease the memory footprint. We also tie token embeddings with weights in the softmax layer~\citep{press-wolf-2017-using} to further reduce  model size. Without these modifications, our 16-layer NPLM does not fit in GPU memory, precluding training.\footnote{Our models are trained on 4 GeForce GTX 1080Ti GPUs.}

\paragraph{Global context representation:} Prior research demonstrates the effectiveness of representing large chunks of text using \emph{averaged} token embeddings~\citep{iyyer-etal-2015-deep,wieting2015towards}. We leverage this work by applying a simple learned kernel (i.e., a 1-D convolution) to the prefix embeddings (beyond just the previous $k$) and including the resulting vector as an extra embedding
to the concatenation layer. We also experiment with replacing the learned kernel with a uniform average. Adding these simple global embeddings improves the NPLM considerably: our 16-layer model's perplexity drops from \textbf{41.9} to \textbf{31.7} with the kernel-derived embedding, while the uniform average achieves a perplexity of \textbf{37.7}.

\section{Using NPLMs to improve Transformers}

While our upgraded NPLM achieves a massive perplexity reduction compared to the original implementation, it is still $\sim 6$ perplexity points short of the baseline Transformer LM. Are there any takeaways from our results that can be used to \emph{improve} Transformer LMs? In this section, we begin with an analysis experiment on \wtthree\ that shows NPLMs outperform Transformers when given shorter prefixes. Inspired by this result, we propose two variants of a Transformer LM that integrate elements of the NPLM, and discover that both of them decrease perplexity across three word-level language modeling datasets (Table \ref{tab:main-res}).

\subsection{NPLMs are better with short contexts} \label{sec:short-context}
Since NPLMs only concatenate a small, fixed number of prefix tokens together, they are obviously unsuited to handle global context. While our upgraded variant addresses this issue to some extent by including aggregated global prefix embeddings into the concatenation layer, the perplexity gap between NPLMs and Transformer LMs remains large. Here, we attempt to understand how much of this difference can be attributed to the Transformer's ability to better model global context. In particular, we train different NPLM and Transformer LMs by truncating the input prefix length to between 3 and 50 tokens. Our NPLM models do not have any global context embeddings in these experiments, and both the NPLM and Transformer models are 16 layers with $\sim$148M parameters each.

\begin{figure} [t]
    \centering \includegraphics[width=0.4\textwidth]{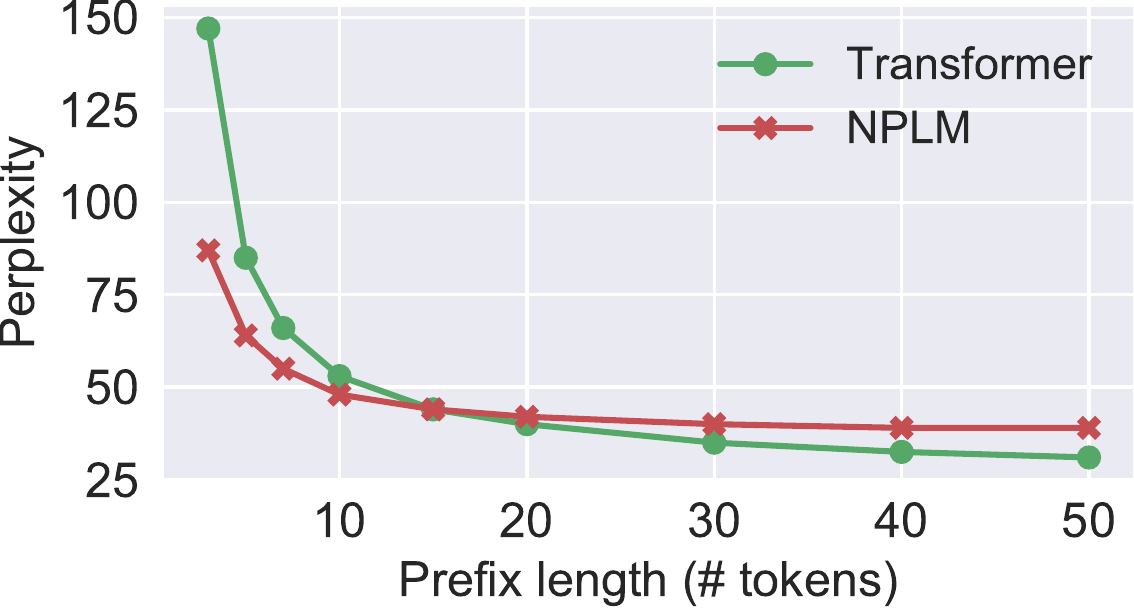} 
    \caption{On the \wtthree\ validation set, NPLM is better than the Transformer with short prefixes but worse on longer ones. 
    }
    \label{fig:short-context-l0-constrain}
\end{figure}

Figure~\ref{fig:short-context-l0-constrain} shows that NPLMs are actually \emph{better} than Transformers when the input sequences are short (i.e., fewer than twenty prefix tokens), but as the prefixes get longer, NPLM perplexity plateaus, while the Transformer perplexity continually decreases. The plot shows that while multi-headed self-attention is effective for longer sequences, it may not be best for modeling shorter contexts.

\subsection{Transformer variants}

Inspired by these results, we investigate hybrid NPLM and Transformer models to better model both short and long-range contexts. In particular, we create two variants of the Transformer by modifying only its \emph{first} layer (L0), while keeping every other layer the same. In the first modification, \textbf{Transformer-N}, we simply replace the first self-attention block in L0 with the NPLM's local concatenation layer (Equation~\ref{eq:concat}), without including any global embeddings. Wondering if the behavior of the concatenation layer can be replicated by self-attention, we also design \textbf{Transformer-C}, in which the self-attention window in L0 is constrained to the previous 5 tokens. 
This constraint is similar to the windowed attention approaches previously applied at all layers in prior Transformer variants~\citep{Beltagy2020Longformer,Roy2020EfficientCS}.\footnote{We do not observe improvements when using local attention at all layers.}

\subsection{Experimental details}

\paragraph{Datasets} We evaluate our models on four language modeling datasets: \wttwo\ and \wtthree~\citep{merity2016pointer}, \lambada~\citep{paperno-etal-2016-lambada}, and the character-level  \enwik\ benchmark~\citep{merity2017regularizing}. 
For \wttwo\ and \wtthree\ ~\cite{merity2016pointer}, we insert an \texttt{<eos>} token after each line, following~\citet{merity2018analysis}. We use adaptive softmax~\cite{pmlr-v70-grave17a} on \wtthree\ with cutoffs $(2e4, 4e4, 2e5)$. On \lambada, we follow~\citet{paperno-etal-2016-lambada} by considering only the most frequent 60K words and replacing the rest with \texttt{<unk>} tokens. We use the preprocessing script released by \citet{merity2017regularizing} to process \enwik.

\paragraph{Models} We train 16-layer (16L) models on the larger \wtthree\ and \lambada\ datasets, 12L models for \enwik, and 6L for the small \wttwo\ dataset.\footnote{The relatively high \wttwo\ perplexities are likely because we did not apply separate regularization that~\citet{merity2017regularizing} show is useful for such a small dataset.} For each dataset, we scale embedding and hidden dimensionality to ensure that all models have roughly the same number of parameters. After tuning hyperparameters on the validation data, we set the number of local concatenated tokens to 15 and the number of 1-D convolution kernels to 5. 

\paragraph{Training details} Our NPLM is trained with dropout probability $p=0.2$, while the other models use $p=0.1$ on all datasets except for \wttwo, for which they use $p=0.3$. For all models, we use the Adam optimizer with $\beta_1 = 0.9$ and $\beta_2 = 0.999$, and training is conducted on 1080Ti GPUs. During evaluation, we follow the methodology of~\cite{knnlm} by providing extra prior context for the scored tokens, 
for instance, in a block of 512 tokens, only the last 128 tokens are scored with the first 384 tokens as context. Detailed architecture, training, and evaluation configurations are included in Appendix~\ref{appendix:config}.


\subsection{Results and analysis}

\begin{figure} [t]
\centering
     \includegraphics[width=0.4\textwidth]{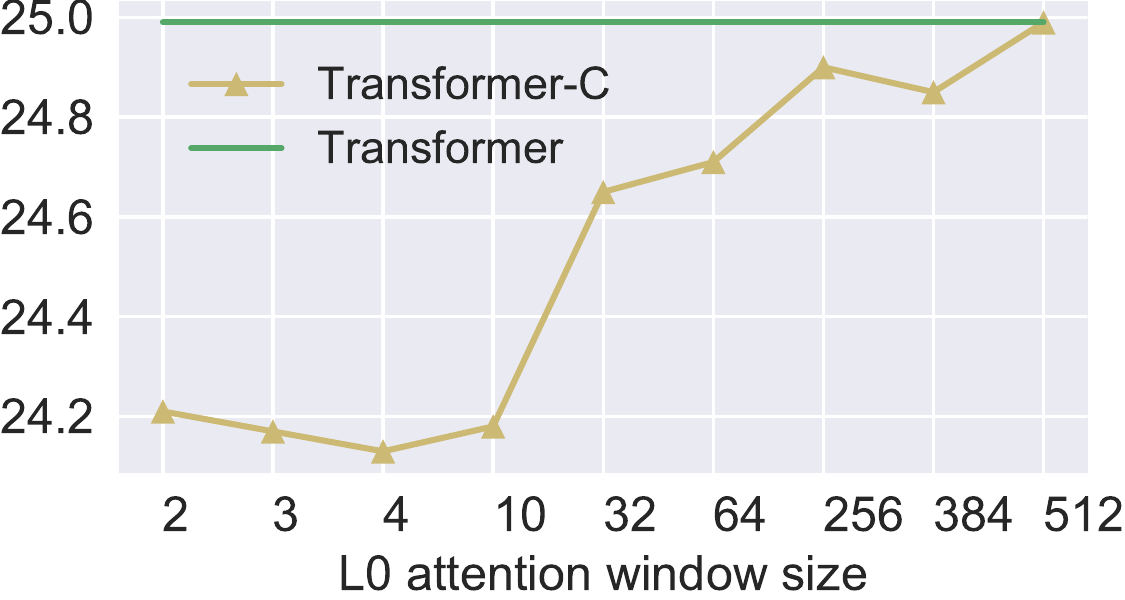}
    \caption{ Transformer-C perplexity decreases with small L0 attention windows.
    }
    \label{fig:short-context-l0-constrain-1}
\end{figure}

Table~\ref{tab:main-res} shows that \textbf{Transformer-N} improves over the baseline Transformer across all three word-level language modeling benchmarks, with the biggest perplexity drop coming on the small \wttwo\ dataset, although character-level perplexity on \enwik\ is unchanged.  \textbf{Transformer-C} also outperforms the baseline Transformer but by smaller margins than \textbf{Transformer-N}. 

\paragraph{Narrower window size in L0 is better:}
We examine \wtthree\ val. perplexity as a function of Transformer-C window size. 
Figure \ref{fig:short-context-l0-constrain-1} shows drops of $\sim$ 1 perplexity point with window sizes of 2-4, which disappear as window size is increased. This experiment supports the importance of focusing on local context at lower layers.

\paragraph{Hybrid models improve at predicting entities and rare words:}
To obtain a more fine-grained understanding of our models, we turn to the long-distance dependency prediction task in \lambada ~\cite{paperno-etal-2016-lambada}, a manually-annotated subset of the full dataset in which correctly predicting a token is possible only when longer contexts are provided.

Table \ref{tab:lambada-acc} shows that our upgraded NPLM achieves less than 1\% accuracy (argmax prediction) on the test set but 30\% on a control set that does not test long-term dependencies. As the baseline Transformer reaches over 30\% accuracy on the test set, this result shows that the convolutional kernels in our modernized NPLM are incompetent at modeling long-range context. 

On the other hand, both \textbf{Transformer-N} and \textbf{Transformer-C} outperform the baseline Transformer (Table \ref{tab:lambada-acc}) by over 1.5\% on the test set. To better understand these improvements, we perform a fine-grained analysis of the tokens for which these models improve over the Transformer. This analysis reveals that the gains stem mainly from three types of target tokens: (1) context-freqeunt (CF) tokens that appear more than twice in the prefix; (2) low frequency tokens (LF) with frequency below 1500; and (3) named entity tokens (Ent) detected by the spaCy~\citep{spacy} NER tagger. The three right-most columns of Table~\ref{tab:lambada-acc} shows that both Transformer variants are more accurate at predicting these tokens, which demonstrates the benefits of enforcing local focus at the first layer.


\begin{table}[t]
    \centering
    \scalebox{0.8}{
        \begin{tabular}{lccccc}
    \toprule
        Model & Test  & Control & CF & LF & Ent.\\\cmidrule(lr){1-3} \cmidrule(lr){4-6}
        NPLM & 0.40  & 30.46 & - & - & -\\
        Transformer & 30.60  & 35.84 & 38.94 & 29.47 & 32.26\\
        Transformer-N & \textbf{32.51}  & 37.06 & 42.33 & 30.14 & 33.95\\
        Transformer-C & 32.23 & \textbf{37.34} & \textbf{42.65} & \textbf{31.58} & \textbf{35.03}\\
    \bottomrule
    \end{tabular}
    }
    \caption{NPLM and Transformer variants on LAMBADA target word accuracy (\%). 
    Variants perform better on context-frequent (CF) tokens that appear at least twice in previous context, low frequency (LF) tokens with frequency < 1500, and named entities (Ent). 
    }
    \label{tab:lambada-acc}
\end{table}

\section{Related work}

The NPLM model in this paper based entirely on the original formulation from~\citet{Bengio2003ANP}.
The variants in our analysis are based on the Transformer model~\citep{NIPS2017_3f5ee243} and Transformer LMs~\cite{baevski2018adaptive, dehghani2018universal,dai-etal-2019-transformer,sukhbaatar-etal-2019-adaptive,knnlm,lm-with-transformers,press-etal-2020-improving,mandava2020pay,press2020shortformer}. 
The constrained local attention in Transformer-C is adopted at all layers of models such as Longformer ~\cite{Beltagy2020Longformer} and Big Bird~\cite{zaheer2020big} due to its sparsity. Our work conceptually resembles that of~\citet{chiu-rush-2020-scaling}, who modernize HMM language models, as well as simple RNN-based language models~\cite{merity2018analysis}. Our linguistic analysis is inspired by experiments from~\citet{khandelwal-etal-2018-sharp}.
\section{Conclusion}
We discover that general-purpose advances in neural architecture design, hardware, and optimization significantly improve the NPLM, a classic language model. An analysis of our upgraded NPLM inspires us to hybridize it with a modern Transformer LM and obtain perplexity decreases across three word-level LM datasets.
\section*{Ethics statement}

\paragraph{Misuse of language models}
Our research involves training large language models on publicly available benchmark datasets. They share the same issues faced by many pretrained language models, such as being used maliciously to generate unfaithful, biased or offensive output.

\paragraph{Energy costs}
We train our models and variants on 4 GeForce GTX 1080 Ti GPUs for all datasets except \wttwo.  We use only one GPU for experiments on \wttwo. The Transformer and its variants take longer to train (40h, 102h, and 108h on \wtthree, \lambada, and \enwik\ respectively). Our modernized NPLM does not have attention module, and therefore trains relatively faster (32h, 45h, and 88h for the above datasets). The energy costs of training and tuning these models, as well as doing exploratory experiments in the initial stages of the project, cannot be ignored. That said, compared to Transformer models, the modernized NPLM has significantly reduced training time, and hence carbon costs. We hope our work contains useful insights for future research that aims to develop simpler and more efficient language models.

\section*{Acknowledgements}
We thank Nader Akoury, Andrew Drozdov, Shufan Wang, and the rest of UMass NLP group for their constructive suggestions on the draft of this paper. We also thank the anonymous reviewers for their helpful comments. This work was supported by award IIS-1955567 from the National Science Foundation (NSF).
\bibliography{anthology,custom}
\bibliographystyle{acl_natbib}

\newpage
\appendix

\section{Experiment details} \label{sec:appendix-exp}


\begin{table} [ht]
    \centering
    \scalebox{0.9}{
    \begin{tabular}{lcc}
    \toprule
        Dataset & Train \#Tokens & Vocab. size  \\\midrule
        \wttwo &  2M & 33K \\
        \wtthree & 103M & 267K \\
        \lambada & 203M & 60K \\
        \enwik & 100M & 205\\
    \bottomrule
    \end{tabular}
    }
    \caption{Dataset statistics}
    \label{tab:datasets}
\end{table}
Dataset statistics are shown in Table \ref{tab:datasets}. 

\begin{table}[ht]
    \centering
   \scalebox{0.85}{
    \begin{tabular}{lccc}
    \toprule
      Dataset   & Train len & Test len & Tgt. len \\ \midrule
      \wttwo   & 512 & 512 & 128 \\
      \wtthree & 512 & 512 & 128 \\
      \lambada & 512 & 512 & 128 \\
      \enwik & 1024 & 1024 & 512 \\
    \bottomrule
    \end{tabular}
   }
    \caption{Training sequence length as well as scored target length and total test sequence length during evaluation we used on each dataset.}
    \label{tab:eval-config}
\end{table}

\begin{table*}[ht]
\centering
\scalebox{0.8}{
\begin{tabular}{@{}lcccccccc@{}}
\hline
\multicolumn{1}{l}{} & \multicolumn{2}{c}{\wttwo} & \multicolumn{2}{c}{\wtthree} & \multicolumn{2}{c}{\enwik} & \multicolumn{2}{c}{\lambada} \\ \cmidrule{2-3} \cmidrule{4-5} \cmidrule{6-7} \cmidrule{8-9}
      & NPLM        & Transformer       & NPLM        & Transformer        & NPLM        & Transformer        & NPLM     & Transformer    \\ \hline
\# Layers     &   6  & 6 &   16 &  16 & 12 & 12 & 16& 16    \\
Emb. dimension   &  256  &  256 & 410 & 410 & 512 & 512 & 512 & 512   \\
Hidden dimension   &  1024 & 1024 &    2100        & 2100      & 2048        & 2048      & 4096       & 4096  \\
Concat hidden dimension & 400 & - &   2000        & -  & 1400        & -  & 2000       & -     \\
\# Attention heads    & -  & 4 &        -  & 10 & -  & 8  & - & 16    \\
Adaptive softmax    & no & no &   yes& yes& no & no & no& no    \\
\# Concat tokens   & 15 & - &   15 & -  & 15 & -  & 15& -     \\
\# Kernel global   &  5 & - &  5  & -  & 5  & -  & 5 & -     \\
Dropout      &     0.3 & 0.3  & 0.2& 0.1& 0.2& 0.1& 0.2        & 0.1   \\\hline
\#Param       &     13M & 13M &    149M        & 148M      & 38M& 38M& 115M       & 115M  \\
\hline
\end{tabular}
}
\caption{Model configuration on \wttwo\ , \wtthree\ , \enwik\ , \lambada\ . }
\label{tab:model-configs}
\end{table*}

\begin{table*}[ht]
    \centering
    \scalebox{0.8}{
    \begin{tabular}{lccccc}
    \toprule
& Warmup steps & Learning rate & Max steps & Batch size & Training time  \\ \midrule
       \wttwo & 100 & 5e-4 & 10k & 5120 & 1.2h/1h\\
       \wtthree & 4k & 2.5e-4/3.5e-4 & 200k & 10240 & 40h/32h\\
       \enwik & 0 & 2.5e-4 & 400k & 22528 & 102h/45h\\
       \lambada & 4k & 3e-4 & 400k & 8192 & 108h/88h\\
      \bottomrule
    \end{tabular}
    }
    \caption{Details of training on the four datasets. Models are trained on single 1080Ti GPU for \wttwo, and on four 1080Ti GPUs for the rest datasets. When a configuration is different for Transformer and NPLM, it's shown in the order Transformer/NPLM.}
    \label{tab:train-config}
\end{table*}



\noindent \textbf{Evaluation} We follow the practice in ~\cite{knnlm} to provide extra prior context for the scored tokens. We provide the training sequence length, test total sequence length, and test target sequence length in Table \ref{tab:eval-config}.

\section{Model configurations} \label{appendix:config}

Detailed model configurations are shown in Table \ref{tab:model-configs}.  Training details are shown in Table \ref{tab:train-config}.

\end{document}